%% file: main.tex
\documentclass[a4paper]{article}

\usepackage{INTERSPEECH2020}

\usepackage{graphicx}
\usepackage{times}
\usepackage{amsfonts}
\usepackage{hyperref,xcolor}
\usepackage{bm}
\usepackage{amsmath}
\usepackage{amssymb}
\usepackage{subcaption}
\usepackage{array,multirow}
\usepackage{latexsym}
\usepackage{xspace}
\usepackage{enumitem}
\usepackage{amssymb}
\usepackage{pifont}

\usepackage{booktabs}
\usepackage{xspace}

\newcommand{\milan}{\textsc{Milan}\xspace}

\newcommand{\flickr}{\textsc{FACC}\xspace}
\newcommand{\conceptual}{\textsc{CSC}\xspace}
\newcommand{\places}{\textsc{PlacesAudio}\xspace}
\newcommand{\loca}{\textsc{LocNarr}\xspace}
\newcommand{\cpc}{\textsc{CPC-8k}\xspace}
\newcommand{\harcnn}{\textsc{Shallow Speech Enc.}\xspace}
\newcommand{\wavtovec}{\textsc{wav2vec}\xspace}

\newcommand\blfootnote[1]{%
  \begingroup
  \renewcommand\thefootnote{}\footnote{#1}%
  \addtocounter{footnote}{-1}%
  \endgroup
}

\title{Talk, Don't Write: A Study of Direct Speech-Based Image Retrieval}
\name{Ramon Sanabria$^{1*}$, Austin Waters$^{2*}$, Jason Baldridge$^2$}
\address{
  $^1$The University of Edinburgh\\
  $^2$Google Research}
\email{r.sanabria@ed.ac.uk, \{austinwaters, jasonbaldridge\}@google.com }

\begin{document}
\ninept
\maketitle

\begin{abstract}

\blfootnote{$^*$Equal contribution.}\blfootnote{$^1$Work done during an internship at Google Research.}Speech-based image retrieval has been studied as a proxy for joint representation learning, usually without emphasis on retrieval itself. As such, it is unclear how well speech-based retrieval can work in practice -- both in an absolute sense and versus alternative strategies that combine automatic speech recognition (ASR) with strong text encoders. In this work, we extensively study and expand choices of encoder architectures, training methodology (including unimodal and multimodal pretraining), and other factors. Our experiments cover different types of speech in three datasets: Flickr Audio, Places Audio, and Localized Narratives. Our best model configuration achieves large gains over state of the art, \textit{e.g.}, pushing recall-at-one from 21.8\% to 33.2\% for Flickr Audio and 27.6\% to 53.4\% for Places Audio. We also show our best speech-based models can match or exceed cascaded ASR-to-text encoding when speech is spontaneous, accented, or otherwise hard to automatically transcribe.

\end{abstract}
\noindent\textbf{Index Terms}: speech-image retrieval, multimodality

\input{00_introduction}
\input{05_related_work}
\input{20_architecture}

\input{40_speech_results}

\input{60_conclusion}

\bibliographystyle{IEEEtran}

\bibliography{mybib}

\end{document}

%% file: 00_introduction.tex
\section{Introduction}
\label{sec:intro}

People speak before they read and write. Humankind used speech exclusively long before the \textit{invention} of writing. People also learn and use language in the real world---to collaborate, describe and relate their visual environment, talk about each other, and more.  While it is hard to capture such environmental grounding broadly, a recent thread of research has focused on the constrained yet rich problem of learning representations of both speech and images using spoken image captions. 

Starting with Harwath and Glass's collection of spoken captions for Flickr8k \cite{harwath2015deep, rashtchian2010collecting}, much research has followed.
A number of works address cognitive and linguistic questions, such as understanding how different learned layers correspond to visual stimuli \cite{gelderloos-chrupala-2016-phonemes,chrupala2017representations}, learning linguistic units \cite{harwath2019learning,harwath2019towards} or how visually grounded representations and data can help understand lexical competition in phonemic processing \cite{havard2019word}.  Other work addresses applied tasks, including multimodal retrieval \cite{ilharco2019large,higy2020textual,rouditchenko2020avlnet}, predicting written keywords given speech and image inputs \cite{kamper-etal-keywords:2017}, cross-modality alignment~\cite{harwath2018jointly}, retrieving speech in different languages using images as a pivot modality~\cite{kamper2018visually,azuh2019towards,ohishi2020trilingual}, and speech-to-speech retrieval \cite{azuh2019towards, ohishi2020trilingual}).

\emph{Speech-based image retrieval} tasks are a dominant evaluation paradigm throughout this body of work. Despite this, it remains difficult to distill lessons about the practicality of speech-based retrieval or how it can be further improved.
One reason is that results have been reported on multiple \textbf{datasets} with different lengths and types of speech. Methodological differences also make comparisons difficult. While many models are variants of dual encoders \cite{gillick2018end} -- a family of architectures well suited for retrieval -- choices of \textbf{speech} and \textbf{image encoders}, \textbf{pretraining} methods, and other factors vary widely.

To address these issues, we run extensive dual encoder experiments over three spoken image-caption datasets -- Flickr Audio \cite{harwath2015deep}, Places Audio \cite{harwath2017learning}, and Localized Narratives \cite{PontTuset_eccv2020} -- which have different properties, including accents, length, and read versus spontaneous speech. Using our implementation, \milan (Multimodal Image and LAnguage Networks), we assess the impact of the choices of encoders, pretraining methods, speech representations, and other factors, and combine best practices to achieve state-of-the-art retrieval performance.


Our primary contributions include:

\begin{itemize}
    \item Showing the \textit{combined} effectiveness of strong pretrained base representations for each modality \cite{schneider2019wav2vec,kawakami2019unsupervised,tan2019efficientnet,bert} with pretraining for the joint speech-image task \cite{ilharco2019large}.
    \item Showing that speech domains impact pretraining: \textit{e.g.}, pretraining on synthetic US English speech and fine-tuning on Localized Narratives (Indian English) degrades performance versus no pretraining.
    \item Obtaining huge gains over state-of-the-art for speech-image retrieval: \textit{e.g.}, \milan bumps speech-to-image R@1 from 21.8\% to 33.2\% for Flickr Audio and 27.6\% to 53.4\% for Places Audio. We also establish the first speech-image retrieval results on Localized Narratives.
    \item Showing for the first time that encoding speech directly can be more effective than using ASR plus a text encoder -- even when using a much stronger text encoder than prior work. ASR only wins for speech of US English speakers reading short written captions (Flickr Audio).
    \item Open sourcing \milan to enable reproducibility $\&$ reuse.
\end{itemize}

%% file: 05_related_work.tex
\section{Datasets}
\label{sec:previous}
\label{sec:data}

Our experiments use three datasets with captions spoken by people and one with synthetic speech. \textit{Flickr Audio Caption Corpus ({\flickr})} \cite{harwath2015deep} has audio captions for Flickr8k.  \textit{Places Audio Caption Corpus ({\places})} \cite{harwath2016unsupervised} has spontaneous spoken captions for Places 205, a dataset with images from 205 scene classes. \textit{Localized Narratives ({\loca})} \cite{PontTuset_eccv2020} has spontaneous spoken descriptions for four image collections (COCO, ADE20K, Flickr30k, Open Images). 

Collectively, these datasets enable comparisons with much prior work and highlight how properties of speech can affect retrieval. Average utterance length in \flickr, 
\places, and \loca is 4, 20, and 40 sec.\ respectively.
\loca contains speech from Indian English speakers;
this allows us to test model generalization across accents. Both \places and \loca include disfluencies and filler words and have more descriptive captions than \flickr. Each dataset has different written forms of its spoken captions: \flickr has original written captions, \places has ASR transcriptions of spoken captions, and \loca has both ASR transcriptions (from Google's Cloud en-IN ASR model) and manual transcriptions created by the speakers themselves. ASR output for \loca has a higher word error rate than \places, so it is particularly interesting for working directly on speech.

\textit{Conceptual Spoken Captions ({\conceptual})} \cite{ilharco2019large} is a spoken version of the Conceptual Captions dataset~\cite{sharma2018conceptual}. Its speech was synthesized by Google's Cloud Speech API2 service, with voices, pitch, and volume varied for each caption to encourage diversity. We use \conceptual \textit{only} for pretraining, \textit{not} for evaluation.

%% file: 20_architecture.tex
\section{Design Choices}
\label{sec:arch}

Early work on speech-image retrieval focused on single words \cite{synnaeve2014learning, harwath2015deep}; later work learned speech representations of full speech utterances with simple models \cite{harwath2016unsupervised,chrupala2017representations}. More recent work suggests that this same architecture can be used for a many tasks. However, only a few have specifically focused on improving retrieval \cite{ilharco2019large,higy2020textual}, and no one has systematically evaluated the effectiveness of different design choices on multiple datasets. 

Dual encoders are a natural approach for learning joint representations of multiple modalities: they perform well and are efficient to train and use for retrieval. In these, an encoder for each modality produces a fixed-length vector representation; the representations from each encoder are combined, often via dot product, and loss functions are defined to encourage positive pairs to have higher scores than negative pairs. We explore many modeling and optimization choices for dual encoders. It is an important approach: it directly benefits from improvements to each individual modality and supports highly scalable nearest neighbor search for retrieval. Early fusion models with interactions between modalities can more powerfully relate two inputs, but they must compute representations and scores for all pairs--which is inefficient for large-scale retrieval. 

\textbf{Input speech representations:} Prior work uses low-level features to represent audio: either Mel Frequency Cepstral Coefficient (MFCC) \cite{ilharco2019large, higy2020textual} or frequency band spectrograms \cite{harwath2017learning,harwath2018jointly}. Recently, self-supervised architectures trained with a contrastive loss showed improvements in English ASR \cite{schneider2019wav2vec} and multiple languages languages\cite{kawakami2019unsupervised}. We consider two of these: \cpc \cite{kawakami2019unsupervised} trained on 8000 hours of multilingual speech, and \wavtovec \cite{schneider2019wav2vec} trained on 960 hours of English read speech.

\textbf{Image encoders:} Prior work has used various image encoders, including VGG16 \cite{harwath2016unsupervised,harwath2018jointly,chrupala2017representations}, ResNet-152 \cite{merkx2019language,higy2020textual}, ResNet50~\cite{harwath2018jointlycv}, and Inception ResNet-V2~\cite{ilharco2019large}. We select the newer EfficientNet \cite{tan2019efficientnet}, which achieves state-of-the-art on ImageNet. Specifically, we use the EfficientNet B4 variant, which has \emph{fewer parameters} than the aforementioned encoders.

\textbf{Audio encoders:} Harwath~\textit{et al.} in \cite{harwath2016unsupervised}--one of the earliest studies of speech-image retrieval--uses a three-layer CNN, where the size of the receptive field of each convolution is set to capture phonemes, syllables and words. This architecture was extended with residual connections \cite{harwath2018jointlycv} and using Inception ResNet-V2 \cite{ilharco2019large}. Recurrent networks, especially Gated Recurrent Units (GRU), have more generally been employed by NLP researchers \cite{chrupala2017representations,chrupala2019symbolic,merkx2019learning,higy2020textual}. We use a ResNet50 CNN to encode, and compare it to a three-layer CNN \cite{harwath2016unsupervised}.

\textbf{Pretraining strategy:} Many have used ImageNet-pretraining \cite{harwath2016unsupervised,harwath2018jointly,chrupala2017representations,higy2020textual,merkx2019language}. Harwath~\textit{et al.}\ \cite{harwath2018jointlycv} pretrain using AudioSet \cite{gemmeke2017audio} and Flickr Natural Sounds \cite{fang2015captions}. Ilharco \textit{et al.}\ \cite{ilharco2019large} use \conceptual to jointly pretrain a dual encoder. We start with an image tower pretrained on ImageNet and jointly pretrain the dual encoder on \conceptual. 

\textbf{Batch Size:} Prior work trained with different small batch sizes, \textit{e.g.}, 48~\cite{ilharco2019large}, 80~\cite{harwath2018jointlycv}, and 128~\cite{harwath2018jointly}. We increase batch size to 1024 with Tensor Processor Unit (TPU) V3 hardware.

\textbf{Training Strategy:} The distance between paired inputs is commonly computed via dot product on the representations produced for each \cite{harwath2016unsupervised}, typically with a triplet loss.
Ilharco~\textit{et al.} use a bidirectional in-batch sampled softmax loss using all in-batch negative samples~\cite{ilharco2019large}. We use this loss without the margin.

%% file: 40_speech_results.tex
\section{Results}
\label{sec:speech_results}

We investigate the impact of best-in-class encoders for each modality and dual encoder training on retrieval.  We compare this to prior work and run a series of ablations.

\label{sec:experimental_settings}

\input{tables/tablefinal_efficient_placesnew}

\textbf{Experimental settings:} Our best \milan speech-image dual encoder uses a ResNet-50 audio encoder with \cpc speech representations, and an EfficientNet-B4 image tower pretrained on ImageNet. We pretrain the dual encoder on {\conceptual} and fine-tune on the target dataset with a batch size of 1024.

\textit{Speech Preprocessing:} We pad or crop all audio to 40 seconds in {\loca}, 20 seconds for {\places}, and 8 seconds for {\flickr} and {\conceptual}. Experiments show these are the optimal length values. We obtain speech representations via offline preprocessing from the \cpc model\footnote{The authors gave us the features and will open source the model.} \cite{kawakami2019unsupervised}.  This model takes a fixed 20s audio window; longer utterances (in {\loca}) are processed as independent 20 second chunks. \cpc's output is a 1024-d representation of each 10ms of audio---notably, the same frame rate as MFCC frontends in most prior work, and common in speech processing applications. 

\input{tables/tablefinal_efficient_facc}

\textit{Image Preprocessing:} During training, images are augmented with random brightening and saturation, and random cropping (to ${\ge}67\%$ of original area). Crops are rescaled to the image encoder's native resolution (380x380 for EfficientNet-B4). We use the public EfficientNet TensorFlow Hub implementation\footnote{\scriptsize https://tfhub.dev/tensorflow/efficientnet/b4/feature-vector/1} which comes with ImageNet pretrained weights.

\textit{Optimization:} Models are trained using Adam with a learning rate of 0.001 and exponential decay of 0.999 every 1000 training steps, $\beta_1{=}0.9$, $\beta_2{=}0.999$ and $\epsilon{=}10^{-8}$. We train on 32-core slices of Cloud TPU V3 pods with effective batch size of 1024 (unless specified otherwise).
Our dual encoders are pretrained on {\conceptual} (Sec. \ref{sec:arch}). To do so, we train the model as described until convergence ($\approx$87k steps), and select the model checkpoint that maximizes held-out retrieval performance.

\textit{Evaluation:} As in prior work, performance is evaluated by retrieving items in one modality (\textit{e.g.},\ images) given a query of the other (\textit{e.g.}, captions). We report Recall@K (R@K): the fraction of times a correct item was found in the top K results.

\textbf{Main results:}
{Table~\ref{tab:finalfacc} compares \milan to prior work on \textbf{\flickr}.
\milan outperforms the previous best \cite{higy2020textual} by a wide margin -- 11.4\% absolute in Speech{$\rightarrow$}Image R@1. Notably, that model (like \cite{chrupala2019symbolic}) exploits additional text supervision, which \textit{doubles} their speech-only model's R@1 of 10.5\%. Such techniques are complementary to ours, so combining them may yield further improvements.
\milan also outperforms \cite{ilharco2019large} -- the previous best trained only on speech and images -- by 19.3\% absolute in Speech{$\rightarrow$}Image R@1. This model, like ours, was pretrained on \conceptual. This suggests improved retrieval can be attributed to our use of better speech representations and stronger image encoder, which improve even more via joint pretraining.
} 

Table~\ref{tab:finalresultsplacesold} compares \milan to prior work on \textbf{\places}. Note: this is for the \textit{original} validation set of \places, which had no held out test split. Compared to the previous best, \milan nearly doubles R@1 for Speech$\rightarrow$Image and \emph{more than} doubles it for Image$\rightarrow$Speech.
\milan outperforms others that incorporate local information with cross-modal fusion~\cite{harwath2018jointly,harwath2018jointlycv} by more than 20\% absolute in R@1. Although our method can be combined with cross-modal fusion, it would not be able to perform retrieval at scale (see Sec.~\ref{sec:arch}). Finally, we compare \milan to \cite{rouditchenko2020avlnet}, a concurrent work using pretraining on images from 1.2M Youtube videos. It shows improvements consistent with our findings on pretraining.

A new split of \places has been created to include held out test material. This addresses the worry that current results (\textit{e.g.}, Table \ref{tab:finalresultsplacesold}) may involve cherry-picking. It also provides separate dev/test-seen and dev/test-unseen splits to allow comparisons for seen (heard) and unseen (unheard) speakers. Table~\ref{tab:finalresultsplacesnew} shows results on the test-seen and test-unseen sets. In both cases, \milan  was optimized on the respective development sets (\textit{i.e.}, dev-seen and dev-unseen). We also obtained results for the ResDAVEnet model \cite{harwath2018jointlycv} (provided by the authors) trained and evaluated on these splits. It includes a better learning rate schedule and other optimization improvements compared to the model in the original paper. Interestingly, ResDAVEnet shows huge gains when using ImageNet pretraining, which is consistent with our findings, broadly.  \milan nevertheless provides large gains over the more complex model. Interestingly both ResDAVEnet and \milan perform better on unseen speakers than seen ones. 

\input{tables/tablefinal_efficient_placesold_compressed}

\input{tables/tableablation_efficient}

Because these new splits support tuning on separate validation split, we hope future work will focus on the new splits. Our results here for both can help anchor such future evaluations.

\input{tables/tableasr_efficient}

\textbf{Ablation study:}
To understand the effect of each modeling choice, we evaluate variants of our best model, shown in Table~\ref{tab:ablationstudy} for {\flickr}, {\places}, and {\loca} (for simplicity, we use only the Flickr30k subset of \loca).  {\cpc $\rightarrow$ \wavtovec} and {\cpc $\rightarrow$ MFCC} respectively swap {\cpc} speech representations with {\wavtovec} \cite{schneider2019wav2vec} features, a similar self-supervised model (trained only on English) and a standard MFCC frontend (128d frames extracted from 20ms windows every 10ms). The remaining variants use MFCC features and apply a series of cumulative ablations. \harcnn replaces the deep ResNet50 audio encoder with a three-layer CNN \cite{harwath2016unsupervised}. \textsc{No pretraining} removes \conceptual pretraining. \textsc{ResNet152} replaces the EfficientNet-B4 image encoder with  ResNet152\footnote{https://tfhub.dev/google/imagenet/resnet\_v1\_152/feature\_vector/4} (larger, but has poorer ImageNet accuracy). Finally, \textsc{Batch size 128} reduces the training batch size from 1024 to 128 to match (approximately) that used in prior work. In other regards, models are trained as described in Sec. \ref{sec:experimental_settings}.

\input{tables/table_vs_prior_compressed}

Table \ref{tab:ablationstudy} shows that changing elements of our best recipe \emph{generally} lowers performance for all datasets (with two notable exceptions discussed below). Of the speech representations, {\cpc} features perform the best uniformly, followed by {\wavtovec} and MFCCs (the worst by a wide margin). Clearly, self-supervised speech representations are very effective.

We see a large improvement by using a bigger speech encoder. The increase of expressivity combined with pretraining on {\conceptual} is crucial for small datasets (\textit{e.g.}, {\flickr}). We observed that when only increasing the amount of parameters of the model without including more data, the effect is diminished. On the contrary, pretraining negatively affects performance for {\loca}\footnote{The quality drop with pretraining disappears when using {\cpc}.} due to the domain gap between synthesized speech using US English text-to-speech models and Indian English speech. These findings indicate it would help to include accent variation to the synthetic speech diversity of \cite{ilharco2019large}.

Finally, in the last two rows from Table~\ref{tab:ablationstudy}, we observe the effect of batch sizes on the different datasets. Interestingly, we see that increasing the batch size harms {\flickr} but helps larger datasets such as {\places} and {\loca}.

\input{50_text_results}

%% file: tables/tablefinal_efficient_placesnew.tex
\begin{table*}

\resizebox{1\textwidth}{!}{
\begin{tabular}{l|rrr|rrr||rrr|rrr}
 & \multicolumn{6}{c||}{{\places} (test-seen)} &  \multicolumn{6}{c}{{\places} (test-unseen)}\\
\multirow{2}{*}{\bf{Model}} & \multicolumn{3}{c|}{\textbf{Speech $\rightarrow$ Image}} & \multicolumn{3}{c||}{\textbf{Image $\rightarrow$ Speech}} & \multicolumn{3}{c|}{\textbf{Speech $\rightarrow$ Image}} & \multicolumn{3}{c}{\textbf{Image $\rightarrow$ Speech}}\\ 
      & R@1 & R@5   & R@10 & R@1  & R@5 & R@10  & R@1 & R@5 & R@10 & R@1 & R@5 & R@10\\
\hline
ResDAVEnet (base) & 22.7 & 49.5 &  62.1 & 18.4 & 47.1  & 59.0 & 26.1 &  52.9 & 63.8  & 21.6 & 47.8 & 58.6\\ 
ResDAVEnet (ImageNet) & 35.2 & 67.5 &  78.0 & 30.4 & 63.1  & 74.1 & 38.3 &  68.5 & 78.8  & 31.2 & 65.0 &  75.4\\ 
\milan & \textbf{58.4} & \textbf{84.6} & \textbf{90.6} & \textbf{53.8} & \textbf{83.4} & \textbf{90.1}  & \textbf{62.1} & \textbf{86.0} & \textbf{90.5} & \textbf{58.2} & \textbf{85.8} & \textbf{90.9}  \\
\end{tabular}}

\caption{Comparison of our best speech-based \milan model with the ResDAVEnet model \cite{harwath2018jointlycv} on {\places}. ResDAVEnet numbers were provided to us by the authors. ResDAVEnet (base) uses a ResNet50 image encoder trained from scratch. ResDAVEnet (ImageNet) is the same but uses ImageNet pretraining for the ResNet50 image encoder.}

\label{tab:finalresultsplacesnew}
\end{table*}

%% file: tables/tablefinal_efficient_facc.tex
\begin{table}
\centering
\begin{tabular}{l|ccc||ccc}
 \multicolumn{7}{c}{\textbf{FACC  (test)}}  \\[1mm] 
\multirow{2}{*}{\bf{Model}} & \multicolumn{3}{c||}{\textbf{Speech $\rightarrow$ Image}} & \multicolumn{3}{c}{\textbf{Image $\rightarrow$ Speech}} \\ 
     & R@1 & R@5   & R@10 & R@1  & R@5 & R@10  \\
\hline
\cite{chrupala2017representations}  & 5.5 & 16.3 &  25.3 & --  & -- & -- \\
\cite{havard-etal-2020-catplayinginthesnow}  & 9.6 & -- & -- & --  & -- & -- \\
\cite{chrupala2019symbolic}$^\dagger$  & -- & -- &  29.6 & -- & --  & --  \\ 
\cite{merkx2019language}  & 12.7 & 34.9 &  48.5 & 16.0  & 42.8 & 56.1 \\ 
\cite{ilharco2019large}  & 13.9 & 36.8 &  49.5 & 18.2  & 43.5 & 55.8 \\
\cite{higy2020textual}$^\dagger$ & 21.8 &  49.9 &  63.1 & --  & -- & -- \\
\milan & \textbf{33.2} & \textbf{62.7} & \textbf{73.9} & \textbf{49.6} & \textbf{79.2} & \textbf{87.5} \\

\end{tabular}

\caption{The best speech-based \milan model compared to previous work on the \flickr test set. $^\dagger$ indicates results from models that use text supervision in a multitask learning objective.}
\label{tab:finalfacc}
\end{table}

%% file: tables/tablefinal_efficient_placesold_compressed.tex
\begin{table}

\centering
\begin{tabular}{l|ccc||ccc}
 \multicolumn{7}{c}{\textbf{\places (val-original)}}  \\[1mm] 
& \multicolumn{3}{c||}{\textbf{Speech $\rightarrow$ Image}} & \multicolumn{3}{c}{\textbf{Image $\rightarrow$ Speech}} \\ 
\bf{Model} & R@1 & R@5   & R@10 & R@1  & R@5 & R@10  \\
\hline
\cite{harwath2016unsupervised}  & 14.8 & 40.3 &  54.8 & 12.1  & 33.5 & 46.3\\
\cite{ohishi2020trilingual}$^\dagger$    & 14.3 &  38.2 & 51.8 & 10.3  & 34.2 & 48.2 \\
\cite{harwath2017learning}$^\ddagger$    & 16.1 &  40.4 &  56.4 & 13.0  & 37.8 & 54.2 \\
\cite{harwath2018jointly}$^\ddagger$ & 20.0 &  46.9 &  60.4 & 12.7  & 37.5 & 52.8 \\
\cite{harwath2018jointlycv}$^\ddagger$  & 27.6 &  58.4 &  71.6 & 21.8  & 55.1 & 69.0 \\
\cite{rouditchenko2020avlnet}$^*$  & 44.8 & 76.9 & \textbf{86.4} & 42.8 & 76.2 & 84.8 \\
\milan  & \textbf{53.4} & \textbf{79.1} & 86.3 & \textbf{53.0} & \textbf{78.2} & \textbf{85.6}
\end{tabular}

\caption{Best speech-based \milan model compared to prior work on \places original validation set. $^\ddagger$ indicates pairwise scoring models that do not support efficient retrieval. $^\dagger$ uses Japanese speech as extra training material. $^*$ concurrent work that uses video data for pretraining.}
\label{tab:finalresultsplacesold}
\end{table}

%% file: tables/tableablation_efficient.tex
\begin{table*}
\centering
\resizebox{\textwidth}{!}{
\begin{tabular}{l|rr|rr||rr|rr||rr|rr}
 & \multicolumn{4}{c||}{{\flickr} (dev) } & \multicolumn{4}{c||}{{\places} (dev-unseen)}  & \multicolumn{4}{c}{{\loca (Flickr30k dev)}}\\
\multirow{2}{*}{\bf{Model}} & \multicolumn{2}{c|}{\textbf{S $\rightarrow$ I}} & \multicolumn{2}{c||}{\textbf{I $\rightarrow$ S}}  & \multicolumn{2}{c|}{\textbf{S $\rightarrow$ I}} & \multicolumn{2}{c||}{\textbf{I $\rightarrow$ S}} & \multicolumn{2}{c|}{\textbf{S $\rightarrow$ I}} & \multicolumn{2}{c}{\textbf{I $\rightarrow$ S}}\\ 
      & R@1 & R@10   & R@1 & R@10 & R@1 & R@10 & R@1 & R@10  & R@1 & R@10 & R@1 & R@10 \\
\hline
Best \milan Model & 37.7 & 75.3  & 52.6 & 87.9 & 64.0 & 92.1  & 61.2 &  92.0  & 70.7  & 95.1  & 68.1 & 95.7 \\
{\cpc $\rightarrow$ \wavtovec}  & 33.3 & 72.2  & 47.7 & 86.5 & 59.2 & 90.0  & 56.7 &  91.4  & 67.5  & 91.2  & 65.0 & 92.0 \\
{\cpc $\rightarrow$ \textsc{MFCC}}  & 20.2 & 55.2  & 33.4 & 75.2 & 51.3 & 86.2  & 47.6 &  85.0  & 60.5  & 88.3  & 58.4 & 89.4 \\
\quad + {\harcnn}  & 11.2 & 39.3  & 16.7 & 54.7 & 37.9 & 73.9  & 31.9 &  70.7  & 29.7  & 63.6 & 25.5 & 60.8 \\
\quad + \textsc{No pretraining}  & 4.1 & 21.0  & 6.3 & 29.9 & 28.0 & 68.1 & 24.6 &  62.1  & 37.4  & 74.9  & 29.6 & 67.7 \\
\quad + \textsc{ResNet152 Image Enc.}  & 1.9 & 10.9  & 2.3 & 16.3 & 18.8 & 53.9  & 19.5 &  54.7  & 22.4  & 54.3  & 20.5 & 50.9 \\
\quad + \textsc{Batch size 128}  & 3.3 & 20.2  & 5.3 & 25.6 & 14.8 & 52.8 & 16.7 & 51.6   & 12.2  & 44.2 & 12.6 & 39.9

\end{tabular}}
\caption{Ablation study for \textit{Speech to Image} (\textbf{S $\rightarrow$ I}) and \textit{Image to Speech} (\textbf{I $\rightarrow$ S}) retrieval on development sets.}
\label{tab:ablationstudy}
\end{table*}

%% file: tables/tableasr_efficient.tex
\begin{table*}
\centering
\resizebox{1\textwidth}{!}{
\begin{tabular}{l|rr|rr||rr|rr||rr|rr}

 & \multicolumn{4}{c||}{{\flickr} (test)}  & \multicolumn{4}{c||}{{\places} (test-unseen)} & \multicolumn{4}{c}{{\loca} (Flickr30k test)}\\ 
 &  \multicolumn{2}{c|}{\textbf{L $\rightarrow$ I}} & \multicolumn{2}{c||}{\textbf{I $\rightarrow$ L}} & \multicolumn{2}{c|}{\textbf{L $\rightarrow$ I}} & \multicolumn{2}{c||}{\textbf{I $\rightarrow$ L}}& \multicolumn{2}{c|}{\textbf{L $\rightarrow$ I}} & \multicolumn{2}{c}{\textbf{I $\rightarrow$ L}}\\

\bf{Language Input}    & R@1 & R@10   & R@1 & R@10  & R@1 & R@10  & R@1 & R@10 & R@1 & R@10 & R@1 & R@10\\
\hline \hline
 
 Speech   & 33.2 & 73.9 & 49.6 & 87.5  & \textbf{62.1} & \textbf{90.5} & 58.2 & \textbf{90.9}  & \textbf{73.5} & \textbf{96.8} & \textbf{74.0} & \textbf{93.5} \\ 

Speech$\rightarrow$ASR$\rightarrow$Text & \textbf{46.9} & \textbf{85.5} & \textbf{63.0} & \textbf{92.8} &  61.1 & 89.6 & \textbf{61.6} & 90.7 & 34.5 & 64.5 &  33.7 & 60.9 \\
\hline

 Ground-Truth Text & 48.0 & 87.7 & 64.1 & 96.3 & --  & -- & -- &  -- & 85.7 & 99.5 & 86.1 & 99.1 \\ 
\end{tabular}}
\caption{Test set results with different language inputs to \milan. Encoding speech directly works a bit better on \places, and R@1 \textbf{doubles} for \loca (which is impacted by poorer ASR). The much higher scores using ground-truth text show there is considerable headroom for better speech encoders and ASR. \textbf{L $\rightarrow$ I} and \textbf{I $\rightarrow$ L} stands for Language to Image, and Image to Language.}
\label{tab:asrresults}
\end{table*}

%% file: tables/table_vs_prior_compressed.tex
\begin{table}
\centering

\begin{tabular}{l|rr|rr}
           &     \multicolumn{2}{c|}{\textbf{Text $\rightarrow$ Image}} & \multicolumn{2}{c}{\textbf{Text $\rightarrow$ Image}}\\
\bf{Model}   & R@1   & R@10  & R@1   & R@10  \\
\hline 
&\multicolumn{4}{c}{\flickr (test)} \\
Chrupala et al. \cite{chrupala2017representations} & 12.7  & 49.4  & --     &  --      \\
Higy et al. \cite{higy2020textual}   & 25.8    &   70.2    & --   & --   \\
Merks and Frank \cite{merkx2019learning} & 27.5 & 70.5  & 38.5     &  79.3 \\
\milan        & \textbf{52.1}  & \textbf{89.5}  & \textbf{65.7}  & \textbf{95.2}\\ 
\hline
&\multicolumn{4}{c}{\places (val-original)} \\
Harwath et al. \cite{harwath2018jointly} & 27.1 & 70.1 & 18.3 & 62.2 \\
\milan        & \textbf{55.2} & \textbf{88.8} & \textbf{58.9} & \textbf{88.9} \\ 
\end{tabular}

\caption{\milan's text encoder performance compared to prior work, showing that it serves as strong competition to retrieval that encodes speech directly. Note that \flickr has ground-truth text from Flickr8k, while \places text is ASR output.}
\label{tab:text_vs_prior_work}
\end{table}

%% file: 50_text_results.tex
\label{sec:text_results}

\textbf{Transcription-based Comparison:} An alternative strategy for speech-to-image retrieval is to use ASR and a modern text encoder (\textit{e.g.}, BERT \cite{bert}), and then train a text tower in a dual encoder using paired text-image data.

Some work with spoken captions has exploited text as an additional training signal \cite{chrupala2019symbolic, higy2020textual}. Some use ground truth text \cite{harwath2018jointly,merkx2019learning}, which is unrealistic for real-world applications. \cite{harwath2018jointly} and \cite{higy2020textual} do compare direct speech encoding to the ASR-text strategy. However, the architectures they used were similar to that for speech. Here, we use much stronger modern text encoding methods to provide a better competitor to direct speech encoders. Where possible, we also compare the performance when using ASR output versus ground-truth written captions.

We create a text tower using an architecture similar to that in \cite{parekh2020crisscrossed}: it computes per-word, contextualized BERT embeddings and inputs these to a 3-layer transformer stack. We pretrain this tower using the written captions in Conceptual Captions \cite{sharma2018conceptual}, which also fine-tunes the ImageNet-pretrained EfficientNet model used for the image tower, similarly to audio pretraining discussed before. Table~\ref{tab:text_vs_prior_work} shows that this text encoder performs far better than previous ones used in the transcribed speech-image retrieval literature---a word-based \cite{chrupala2017representations} and character-based \cite{merkx2019learning} GRU encoder, and a CNN \cite{harwath2018jointly}. It is also competitive with other dual encoders used in text-image retrieval work---it obtains R@1 of 52.0\% for image-to-text and 37.9\% for text-to-image retrieval on MS-COCO 5k \cite{karpathy2015deep}, compared to 53.0\% / 40.5\% for VSRN \cite{Li_2019_ICCV}.

The key question is whether models that encode speech directly can outperform models using such a text encoder on ASR output of the speech. Table~\ref{tab:asrresults} shows indeed this is possible: the direct speech encoder has double the performance of the ASR strategy for \loca. Recall that the speech in \loca is Indian English, and the ASR quality is lower than it is for US English. The direct speech encoder matches the ASR strategy for \places (which has US English speech). 

Compared to using ground-truth transcriptions (possible only for \flickr and \loca), there is clearly much room for improving direct audio encoding versus working with clean, high-fidelity text. For real-world scenarios where speech is spontaneous (\places) or spontaneous and accented (\loca), a strong speech encoder like \milan's is a better option. However, if one has high quality audio and standard, non-conversational speech,  the ASR approach works better.

%% file: 60_conclusion.tex
\section{Conclusions and Future Work}
\label{sec:conclusion}

Our results clarify both machine learning and engineering considerations for learning representations of speech and images from spoken captions. Our dual encoder models learn stand-alone encoders trained jointly and can be used for efficient multimodal retrieval (unlike early fusion models with cross-modal interactions between subparts of audio and images \cite{harwath2018jointly}). Exploring direct representations of speech could free us of ASR errors and reduce dependence on high-quality ASR systems, which are resource intensive and not available for all languages.

Notably, self-supervised speech representations work better than traditional filterbank features, and are free---in the sense that they require no extra human-curated supervision. In this vein, we expect that the recent spate of self-supervised representation learning techniques for images, such as SimCLR \cite{chen2020simple}, are worth investigating as substitutes to ImageNet training.